\documentclass[sn-mathphys-num]{sn-jnl}

\usepackage{graphicx}%
\usepackage{multirow}%
\usepackage{amsmath,amssymb,amsfonts}%
\usepackage{amsthm}%
\usepackage{mathrsfs}%
\usepackage[title]{appendix}%
\usepackage{xcolor}%
\usepackage{textcomp}%
\usepackage{manyfoot}%
\usepackage{booktabs}%
\usepackage{listings}%
\usepackage{booktabs}
\usepackage{makecell}
\usepackage{threeparttable}
\usepackage{algorithmic}

\usepackage{subfigure}

\usepackage[lined,boxed,linesnumbered,ruled]{algorithm2e}

\begin{document}

\title[Article Title]{Predictive Risk Analysis and Safe Trajectory Planning for Intelligent and Connected Vehicles}

\author[1]{\fnm{Han} \sur{Zeyu}}

\author*[2]{\fnm{Cai} \sur{Mengchi}}\email{caimengchi@tsinghua.edu.cn}
\author[1]{\fnm{Chen} \sur{Chaoyi}}
\author[1]{\fnm{Meng} \sur{Qingwen}}
\author[1]{\fnm{Wang} \sur{Guangwei}}
\author[1]{\fnm{Liu} \sur{Ying}}
\author[1]{\fnm{Xu} \sur{Qing}}
\author[1]{\fnm{Wang} \sur{Jianqiang}}
\author[1]{\fnm{Li} \sur{Keqiang}}

\affil[1]{\orgdiv{School of Vehicle and Mobility}, \orgname{Tsinghua University}, \orgaddress{\postcode{100084}, \state{Beijing}, \country{China}}}
\affil[2]{\orgdiv{Department of Civil Engineering}, \orgname{Tsinghua University}, \orgaddress{\postcode{100084}, \state{Beijing}, \country{China}}}

\abstract{The safe trajectory planning of intelligent and connected vehicles is a key component in autonomous driving technology. Modeling the environment risk information by field is a promising and effective approach for safe trajectory planning. However, existing risk assessment theories only analyze the risk by current information, ignoring future prediction. This paper proposes a predictive risk analysis and safe trajectory planning framework for intelligent and connected vehicles. This framework first predicts future trajectories of objects by a local risk-aware algorithm, following with a spatiotemporal-discretised predictive risk analysis using the prediction results. Then the safe trajectory is generated based on the predictive risk analysis. Finally, simulation and vehicle experiments confirm the efficacy and real-time practicability of our approach.}


\maketitle

\section{Introduction}\label{Sec:Introduction}
The intelligent and connected vehicles(ICVs), aiming at providing safe, comfortable and efficient transportation experience for users, have drawn increasing research interest recently \cite{wang2021towards}.  These vehicles leverage advanced technologies such as artificial intelligence, machine learning, and real-time data communication to enhance their operational capabilities. Among all the research domains, studies focusing on the safety of ICVs hold the utmost significance, as safety is the fundamental necessity in transportation \cite{wang2023decision}. Ensuring the safety of ICVs not only protects passengers and pedestrians but also fosters public trust in autonomous driving technologies. Consequently, researchers are exploring various safety measures, including robust sensor systems, fail-safe mechanisms, and comprehensive risk assessment frameworks, to mitigate potential hazards associated with ICV operation.

To guarantee the safety of ICVs, numerous risk assessment algorithms have emerged \cite{yang2025recent, li2020threat}. The initial category of risk assessment algorithm calculates the driving risk based on time, such as time headway (THW) \cite{branston1976models, li2022variable} and time to collision (TTC) \cite{lee1976theory, nadimi2020evaluation}. Further research integrates the kinematic parameters \cite{chang2009rear, wang2023trajectory, shi2024collaborative} or statistic information \cite{greene2011efficient, chen2024spectrum, liu2025comparison} to analysis risk.

Another group of risk assessment algorithm models the driving risk by the risk field. With proper design, the risk field is capable of measuring the risk of the whole scene leveraging situation information. Krogh et al. \cite{krogh1984generalized} first propose the artificial risk field for robot obstacle avoidance in unstructured scenarios. Then this expressive modeling algorithm has been extended to other fields, including laen-keeping assistance system \cite{rossetter2006lyapunov}, traffic flow modeling\cite{ni2011unified}, and lateral trajectory planning \cite{matsumi2013autonomous}. For structured traffic scenarios, Wang et al. \cite{wang2015driving} first construct a unified driving safety field architecture considering the properties and behaviours of traffic participants and road conditions such as road slope and traffic lights.

Most of existing risk assessment theories solely analyze the driving risk by only the current perception information, without considering the predictive information of the surrounding objects in the future. Consequently, these theories fail to reflect the dynamic changes of the environment over time, resulting in relatively conservative decision-making of ICVs.

However, most existing risk assessment theories primarily analyze driving risk based solely on current perception information, without adequately considering the predictive information of surrounding objects in the future. This limitation results in a failure to reflect the dynamic changes of the environment over time, leading to relatively conservative decision-making in ICVs. By not incorporating predictive analytics, these theories may miss critical insights that could inform more proactive and adaptive safety measures.

Therefore, there is a pressing need for the development of risk assessment frameworks that integrate both current and predictive information, enabling ICVs to make more informed and timely decisions in complex driving scenarios. In response to the problems with existing methods, this paper proposes an innovative predictive risk assessment and safe trajectory planning framework for ICVs. The innovations of this paper are as follows.
\begin{enumerate}
    \item First, a local risk-based vehicle trajectory prediction method is proposed. This method enhances the accuracy of predicting surrounding vehicle trajectories by analyzing risks near each object. 
    \item Second, a predictive spatiotemporal-discretized risk analysis method is introduced, which adds a temporal dimension to risk assessments, allowing for a more accurate depiction of the evolution of micro-risks in the traffic environment around the vehicle over time. 
    \item Third, a vehicle safety trajectory planning method based on spatiotemporal-discretized predictive risk assessment is developed. Using the results of the predictive risk assessment, the method performs safety checks on alternative trajectories and ultimately outputs a trajectory that complies with safety constraints. 
    \item Finally, the validity of the methods proposed in this paper is verified using simulation and real-world vehicular driving data.
\end{enumerate}
The remainder of this paper is organized as follows: Section \ref{Sec:Technical} briefly presents the technical framework of the approach proposed in this paper. Section \ref{Sec:Method} describes the detailed algorithms of each module. Simulation and vehicle experiments are conducted in Section \ref{Sec:Results}, and Section \ref{Sec:Conclusion} draws the conclusion of this paper.

\section{Technical Framework}\label{Sec:Technical}

\begin{figure}[htbp]
\centering
\includegraphics[width=0.9\textwidth]{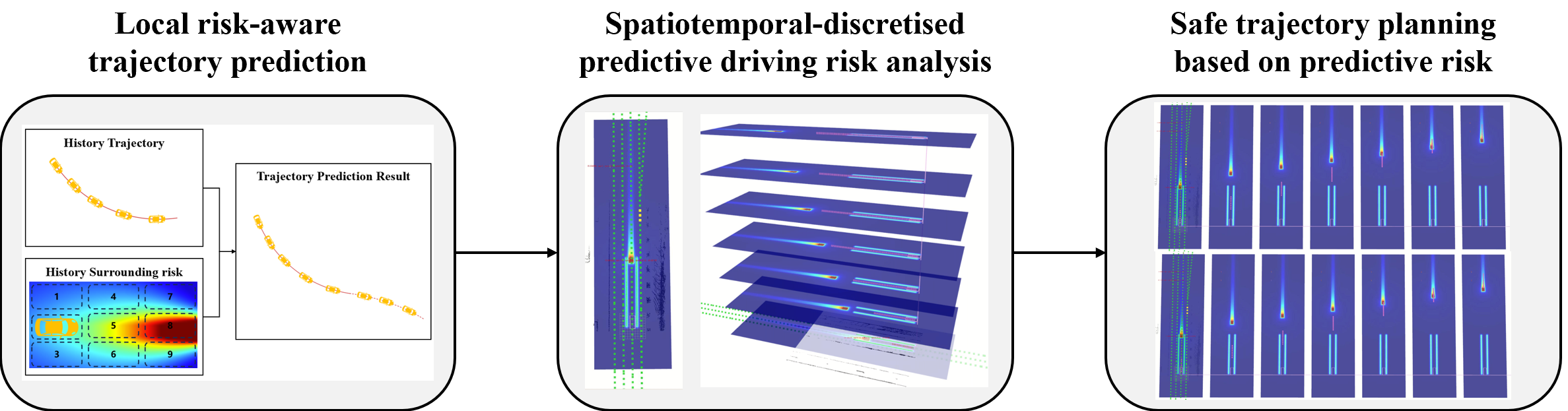}
\caption{Overall architecture of predictive risk analysis and safe trajectory planning}
\label{Fig:Framework}
\end{figure}

The overall architecture of our predictive risk analysis and safe trajectory planning approach is demonstrated in Fig. \ref{Fig:Framework}, consisting of the following three modules:
\begin{enumerate}
    \item Local risk-aware trajectory prediction: based on vehicle sensors, information about surrounding vehicles, pedestrians, and other targets is obtained and retained for a period. This target information is then coupled with map data to more accurately analyze the behavioral intentions of the targets. Models for predicting the behavior and trajectories of vehicles and pedestrians are constructed to enable the prediction of future trajectories of nearby vehicles and pedestrians. The prediction results for the targets include future positional/speed/and other information at various time intervals.
    \item Spatiotemporal-discretised predictive driving risk analysis: based on the results of trajectory prediction of surrounding targets, a spatiotemporal discrete risk field model is constructed. Several risk fields for current and future time intervals are generated according to the perception and prediction information. The risk fields contain potential fields and kinetic fields. The faster the target's speed, the larger the kinetic field in front of it, and the closer the distance to the target, the greater the potential field.
    \item Safe trajectory planning based on predictive risk: utilizing the information from the future time interval risk fields, a safe trajectory planning model is constructed to generate the optimal safe trajectory. Finally, the trajectory is sent to the control module. The vehicle can complete the planning of a safe trajectory based on the information from the risk fields, without perception and prediction information as input.
\end{enumerate}

\section{Methodologies}\label{Sec:Method}
In this section, we will detailedly introduce the specific methodologies of trajectory prediction, risk assessment and trajectory planning in the framework designed in Sec. \ref{Sec:Technical}.

\subsection{Local risk-aware trajectory prediction}

\begin{figure}[htbp]
\centering
\includegraphics[width=0.9\textwidth]{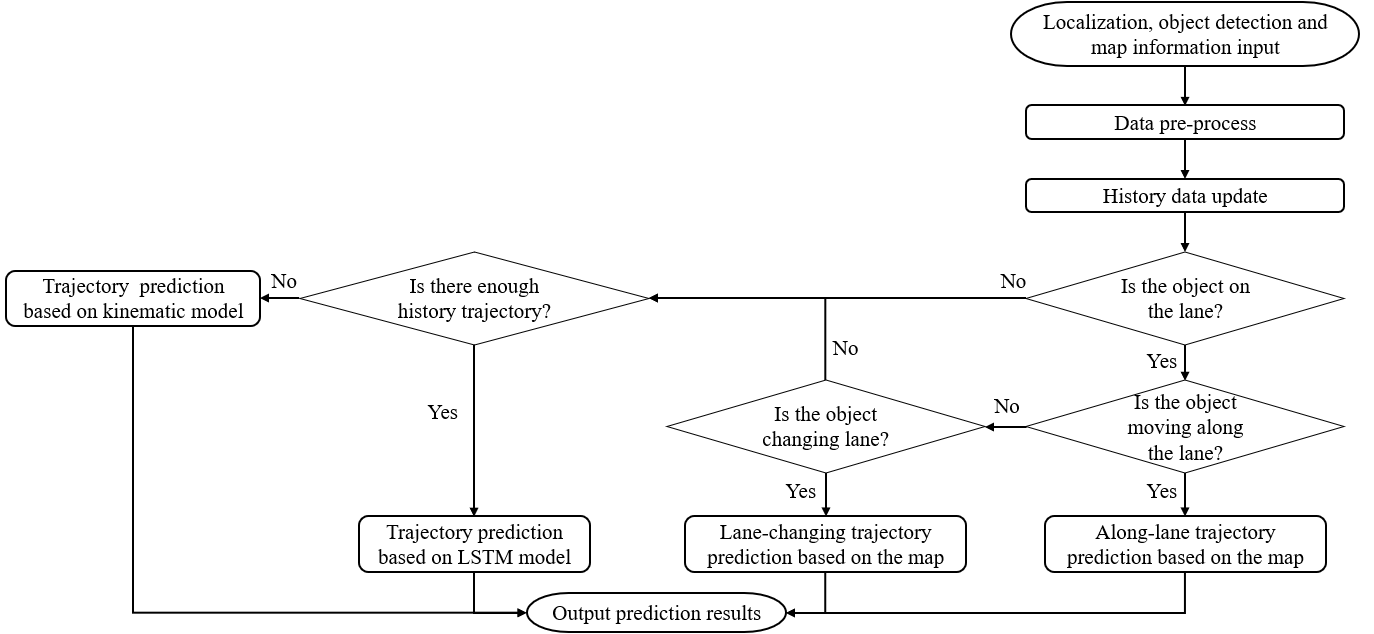}
\caption{The risk-aware trajectory prediction framework}
\label{Fig:Prediction}
\end{figure}

In this subsection, we propose a risk-aware trajectory prediction framework as Fig.\ref{Fig:Prediction} shows. Given that current learning-based trajectory prediction algorithms require a period of history trajectory information as input \cite{liu2021survey}, we incorporate kinematic model and map-based trajectory prediction with Long Short-Term Memory (LSTM)-based trajectory prediction to provide consistent and robust trajectory prediction. When the object to be predicted is not on the lane and has enough history trajectory information, the LSTM model is applied to predict its future trajectory. Otherwise, the kinematic model and map information are utilized to fill in the gaps of learning-based prediction. 

\begin{figure}[htbp]
\centering
\includegraphics[width=0.9\textwidth]{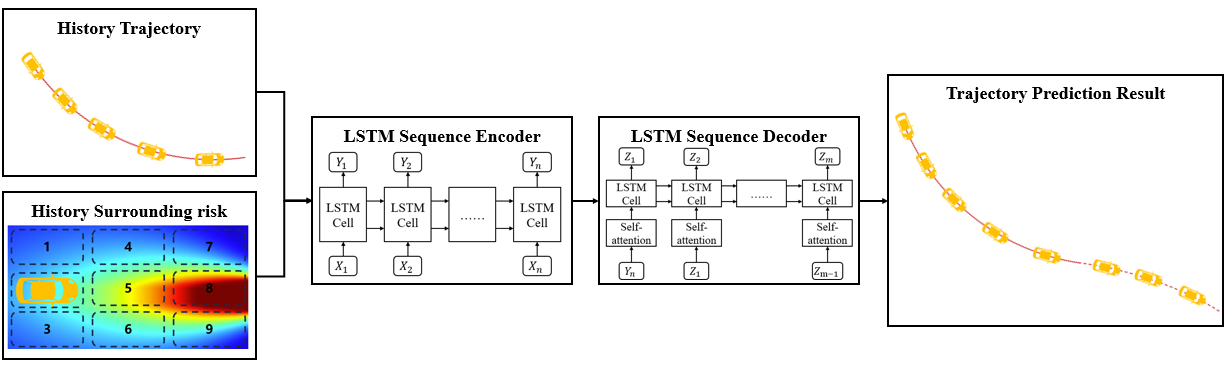}
\caption{The LSTM model with risk assessment and self-attention mechanism.}
\label{Fig:LSTM}
\end{figure}

Specifically, the risk assessment is integrated in the LSTM prediction model to improve its performance. As Fig. \ref{Fig:LSTM} shows, a sequence-to-sequence LSTM model \cite{park2018sequence} with self-attention mechanism \cite{shaw2018self} is established as the backbone model, and each object’s history trajectory is concatenated with its history surrounding risk as the LSTM sequence encoder input $\{X_i\},i=1,2,...,n$ . $X_i$ indicates the $i$th frame of history information, including 2-dimension positions and 9-dimension surrounding risks. The risk of each surrounding position is calculated by other objects refer to \cite{wang2016driving, wang2023sa}. The LSTM sequence decoder is aided by self-attention mechanism to extract weights of input information, and generate the trajectory prediction $\{Z_i\},i=1,2,...,m$. $Z_i$ indicates the $i$th frame of prediction trajectory and contains 2-dimension positions.

\subsection{Spatiotemporal-discretised predictive driving risk analysis}
Most existing driving risk analysis algorithms\cite{} calculate the driving risk utilizing only current environment information, without considering future moments.. To adequately analyze the future driving risk as the guidance of trajectory planning, we propose a spatiotemporal-discretised predictive driving risk analysis algorithm incorporating the trajectory prediction information.

In our approach, we first calculate the risk field at the current moment. Subsequently, based on the predicted positions of each object, we compute the risk field at future moments at intervals of 0.5 seconds. 

Objects of the whole scene are separated into dynamic objects, such as vehicles and pedestrians, and static objects like lane lines and road barriers. The risk field energy of each dynamic object $i$ at each position $(x,y)$ is denoted as $E_d^i (x,y)$, consisting of potential field energy $^pE_d^i (x,y)$ and kinetic field energy $^kE_d^i (x,y)$ as follows:

\begin{equation}
    E_d^i (x,y) = {^p}E_d^i (x,y) + {^k}E_d^i (x,y), i = 1,2,...n.
\end{equation}

The potential field energy $^pE_d^i (x,y)$ describes the potential energy at each position $(x,y)$ of each dynamic object, depending on the mass $m$, velocity $v$ of the object and the distance $D(x,y)$ to the object. We define the virtual mass $M_v^i$ of the object to construct $^pE_d^i (x,y)$:

\begin{equation}
\begin{aligned}
    ^pE_d^i (x,y) &= \frac{k \cdot r_a \cdot M_v^i}{(D^{k_1}(x,y) + k \cdot r_a \cdot M_v^i /E_{max})},\\
    M_v^i &= m \cdot T \cdot (\alpha \cdot v^{\beta}+\gamma),\\
    D(x,y) & =
\begin{cases}
0   & \quad\text{if $(x,y)$ is inside the object,}\\
min\{D_k(x,y)\} &\quad  \text{if $(x,y)$ is outside the object.}
\end{cases}
\end{aligned}
\end{equation}
In the above equations, $k, r_a, k_1, E_{max}, T, \alpha, \beta, \gamma $  are the parameters of the potential field energy, and $D_k(x,y)$ refers to the distance of position $(x,y)$ to the $k$ th edge of the object.

Then we can calculate the kinetic field energy $^kE_d^i (x,y)$ based on the velocity $v$ and speed direction $\delta$ of the object $i$. $\delta$ can be obtained using the trajectory prediction results. The lateral and longitudinal distance of position $(x,y)$ to the object $i$ in its Frenet coordinate system are denoted as $D_{la}(x,y)$ and  $D_{lo}(x,y)$, respectively. To ensure the kinetic field energy decrease along the lateral and longitudinal directions, we propose a weighted distance $D_w(x,y)$ as follows:
\begin{equation}
    D_w(x,y) = \sqrt{\frac{D_{lo}(x,y)^2}{1-(D_{la}(x,y)/w)^2}},
\end{equation}
where $w$ is the width of the object.

In the meanwhile, to characterize the impact of relative velocity $v_r$ of the object to the ego-vehicle, the kinetic field energy $^kE_d^i (x,y)$ is formulated as:
\begin{equation}
    {^k}E_d^i (x,y) = \frac{v_r^2}{2M_v^i \cdot D_w(x,y) \cdot (1-D_w(x,y)/(v_r \cdot t))},
\end{equation}
where $t$ is the time threshold. We also apply the potential energy and logarithmic smoothing to restrict and smooth the kinetic field energy.

\begin{figure}[htbp]
\centering
\includegraphics[width=0.7\textwidth]{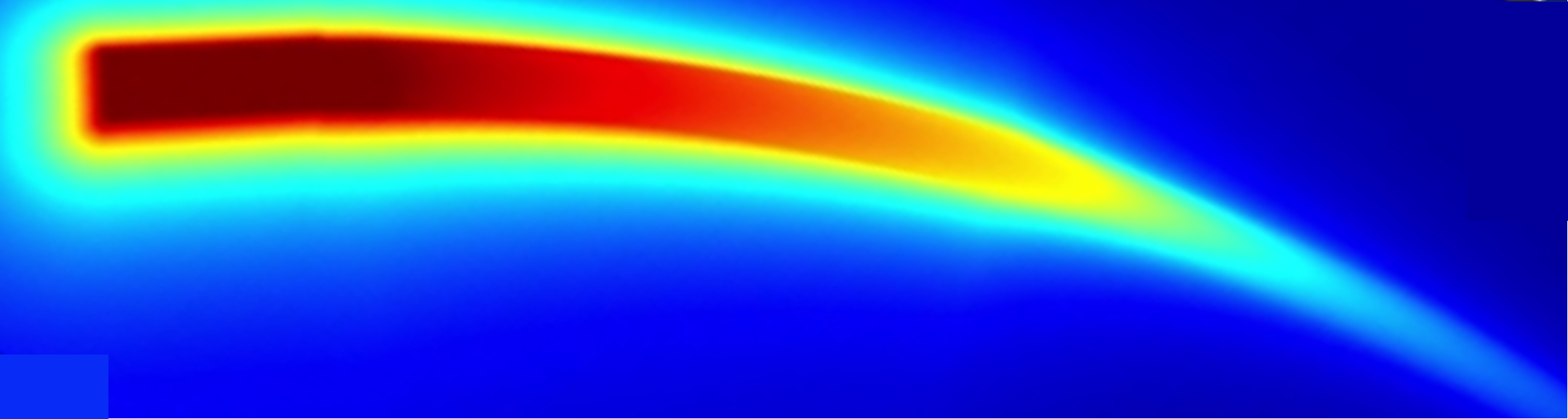}
\caption{The risk field of a turning object.}
\label{Fig:Turning}
\end{figure}

For a turning object, its combined risk field is as Fig. \ref{Fig:Turning} shows. 

As for the static objects, their risk field energy only contains potential field energy as they are stationary. The risk field energy $E_s^j (x,y)$ of the $j$ th static object at position $(x,y)$ can be computed as:
\begin{equation}
    E_s^j (x,y) = k_s \cdot (\frac{\kappa D}{2}-min\{\frac{\kappa D}{2}, dis(x,y)\})^2,
\end{equation}
where $D$ is the width of the static object, i.e. the width of the lane line in most cases, $\kappa\geq 1$ is the ratio parameter, and $dis(x,y)$ is the distance of the position $(x,y)$ to the static object. $k_s$ is a parameter determined by the type of the static object. For instance, $k_s$ of the lane lines are relatively low as they only serve as soft restrictions, while $k_s$ of road barriers can be set to a high enough value to prevent ego-vehicle from colliding with them.

\begin{figure}[htbp]
\centering
\rotatebox{90}{\includegraphics[width=0.3\textwidth]{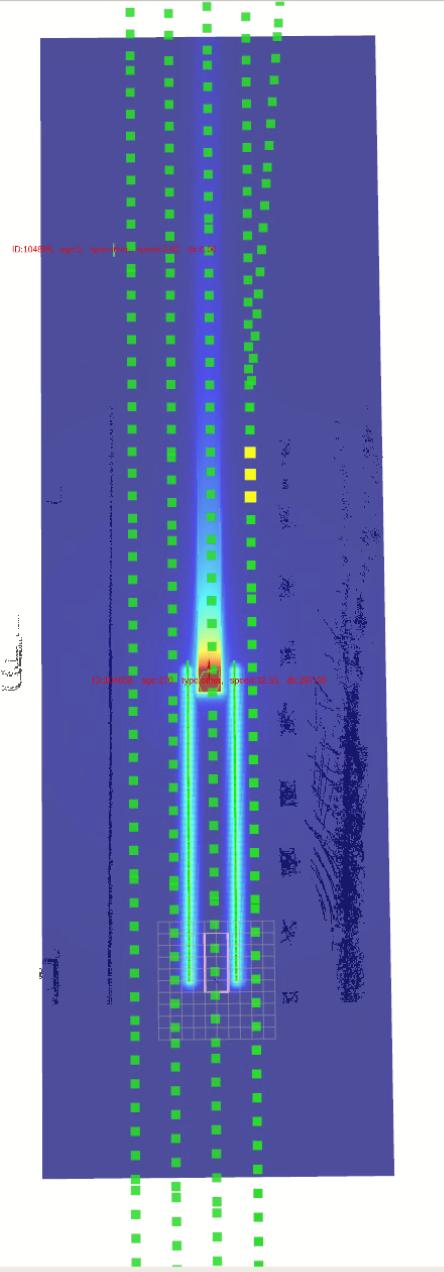}}
\caption{The whole risk field of a dynamic object with static lane lines in a single frame}
\label{Fig:Single}
\end{figure}

\begin{figure}[htbp]
\centering
\includegraphics[width=0.9\textwidth]{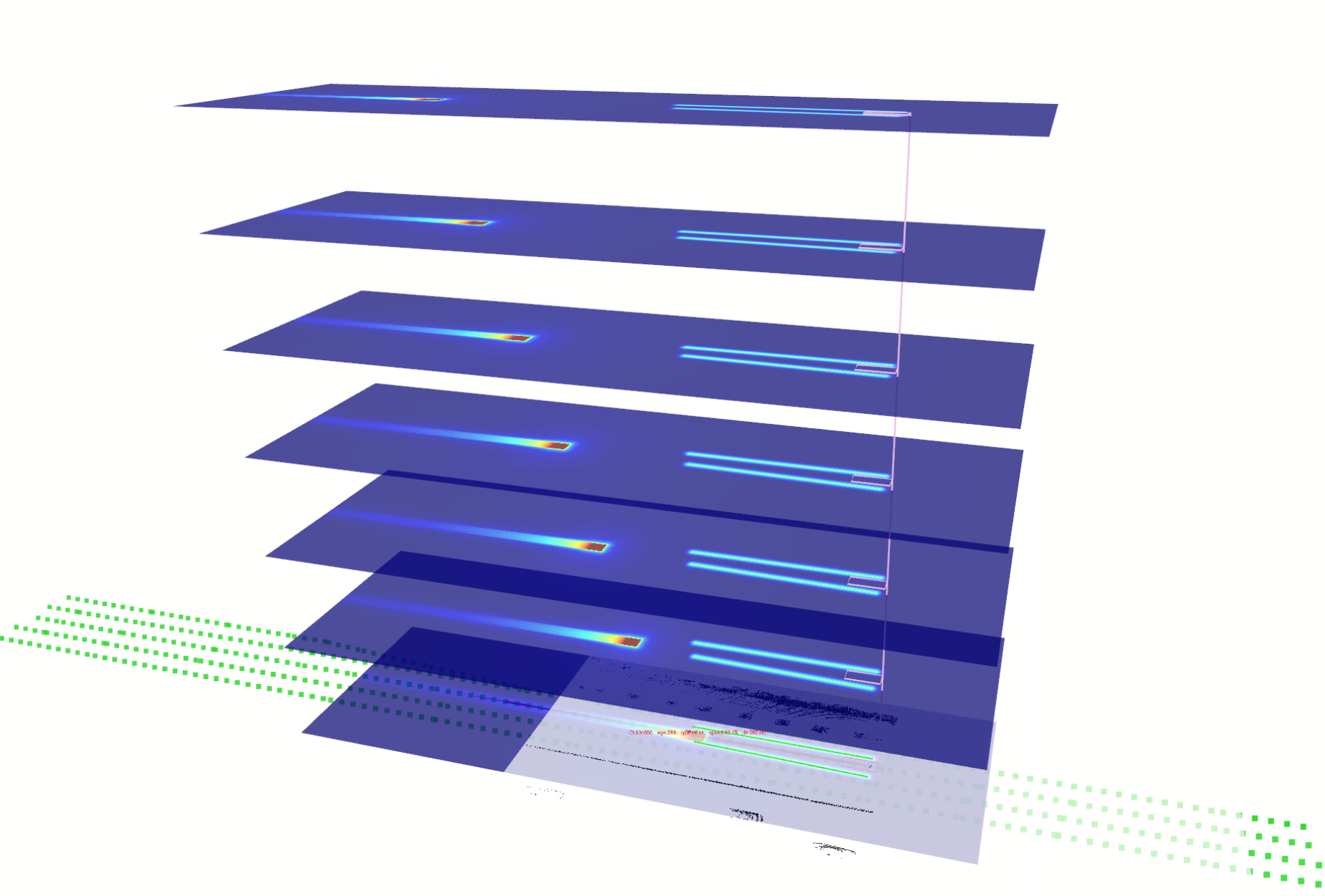}
\caption{The comprehensive spatiotemporal-discretised predictive driving risk field}
\label{Fig:Multi}
\end{figure}

Fig. \ref{Fig:Single} demonstrates the whole risk field of a dynamic object with static lane lines in a single frame.
The future risk fields are calculated in a similar manner based on the trajectory prediction results. Fig. \ref{Fig:Multi} is an example of the comprehensive spatiotemporal-discretised predictive driving risk field.

\subsection{Safe trajectory planning based on predictive risk}
After we generate multiple spatiotemporal-discretised predictive driving risk field matrices $\{M_j\}_{j=1}^m$, they are employed for the safe trajectory planning as Algorithm \ref{Alg:Planning} illustrates. 

\begin{figure}[htbp]
\centering
\includegraphics[width=0.9\textwidth]{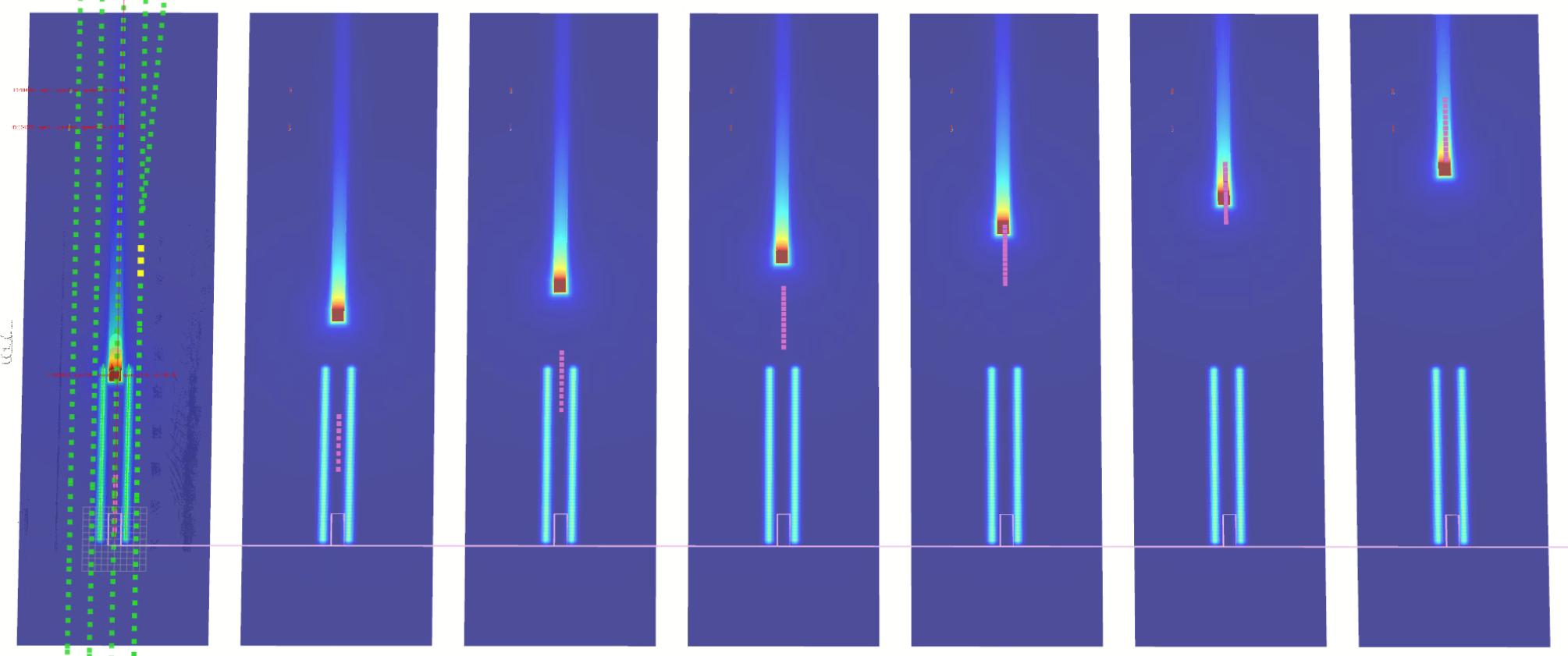}
\caption{The slices of paths and their corresponding risk fields}
\label{Fig:All}
\end{figure}

First, multiple candidate paths $\{Path_i\}_{i=1}^n$ are sampled as Bezier curves in front of the vehicle along the road. Based on the discrete interval timing of the target prediction results and combined with the vehicle's current speed, each path $Path_i$ is sliced into $\{Slice_j\}_{j=1}^m$ for processing. Each path slice considers only the risk field corresponding to that slice. Taking the candidate curve shown in Fig. \ref{Fig:All} as an example, each curve is divided into 7 slices as the red lines shows. Each slice only considers the risk field information around it. The the maximum risk value $r_{max}$ in each slice is utilized for calculating the desired speed $s_j$ for that slice. The desired speed that is the lowest among all the slices is selected as the final desired speed for this candidate path. Finally, the candidate path with the highest desired speed is selected as the final planning path and sent to the control module. the selected trajectory $Traj$ is generated by the highest desired speed $S_{i^*}$ and its corresponding path $Path_{i^*}$.

Slicing the candidate paths not only reflects the spatiotemporal handling of dynamic targets but also avoids calculating the risk field for different future spacetimes along the entire path, ensuring computational efficiency while maintaining rationality.

\begin{algorithm}[t]
    \caption{The pipeline of safe trajectory planning}
    \label{Alg:Planning}
    \KwIn{current position $P$, speed $S$, risk matrices $\{M_j\}_{j=1}^m$, map information $Map$, global target $T$, speed limit $S_{max}$.}
    \KwOut{selected trajectory $Traj$.}
    \SetKwFunction{PathSampling}{PathSampling}
    \SetKwFunction{PathSlicing}{PathSlicing}
    \SetKwFunction{MaxRiskSearching}{MaxRiskSearching}
    \SetKwFunction{SpeedCalculating}{SpeedCalculating}
    \SetKwFunction{TrajectoryGenerating}{TrajectoryGenerating}
    \BlankLine
     Initialize: path desired speeds $\{S_i\}_{i=1}^n$;
     
    Sampled trajectories $\{Path_i\}_{i=1}^n$ = \PathSampling$(P, Map, T)$;

    \For{$Path_i$ in $\{Path_i\}_{i=1}^n$}{
    
        Initialize: slice desired speeds $\{s_j\}_{j=1}^m$;
        
       Path slices $\{Slice_j\}_{j=1}^m$ = \PathSlicing$(S, \{Path_i\}_{i=1}^n)$;
        
        \For{$Slice_j$ in $\{Slice_j\}_{j=1}^m$}{
            Max risk $r_{max}$ = \MaxRiskSearching($Slice_j, M_j$);
            
           Slice desired speed $s_j$ = $\min{ \{ \text{\SpeedCalculating}(r_{max}), S_{max} \}}$;
        }

        Trajectory desired speed $S_i$ =  $\min \{s_j\}_{j=1}^m$
    }

    Selected path index $i^*$ = $argmax\{S_i\}_{i=1}^n$;

    Selected trajectory $Traj$ =\TrajectoryGenerating($Path_{i^*}, S_{i^*}$);

    \KwRet $Traj$
\end{algorithm}

\section{Simulation and Experiments}\label{Sec:Results}
In this section, we conduct abundant simulation and vehicle experiments to validate the proposed predictive risk analysis and safe trajectory planning algorithm.
\subsection{Simulation Results}
Before vehicle experiments, we first finish several simulations to test the effectiveness of trajectory prediction, risk analysis and safe trajectory planning, respectively.

\subsubsection{Trajectory Prediction}
To test the trajectory prediction performance, the LSTM model with risk assessment and self-attention mechanism is trained and tested on a dataset collected in complicated traffic scenarios in Xiongan and Wuhan, China. 30 frames of history information are taken as input and predict the future trajectory of 20, 40, and 60 frames, which are 2, 4, and 6 seconds, respectively. As comparisons, the model without risk assessment and without self-attention mechanism are also trained to validate the effectiveness of these modules. The Average Displace Error (ADE) and Final Displace Error (FDE) results are shown in Tab. \ref{Tab:prediction}.

\begin{table}[!htbp]
    \centering
    \caption{The trajectory prediction results in ADE/FDE(m).}
    \begin{tabular}{cccc}
        \toprule
        Algorithms &\makecell[c]{20 frames \\ prediction} &\makecell[c]{40 frames \\ prediction} &\makecell[c]{60 frames \\ prediction}\\
        \midrule
        Original Model &\textbf{0.49/0.90} &\textbf{0.82/1.84} &\textbf{1.12/2.63} \\
        Without Risk Assessment &0.53/0.97 &0.85/1.89 &1.21/2.76 \\
        Without Self-attention Mechanism &0.62/1.05 &0.84/1.85 &1.23/2.78 \\
        \bottomrule
    \end{tabular}
    \label{Tab:prediction}
\end{table}

As the above results illustrates, the incorporated risk assessment module and self-attention mechanism both observably improve the prediction performance on 20, 40, and 60 frames prediction. The results indicate that the trajectory prediction algorithm can serve as the solid foundation of further modules.

\subsubsection{Risk Analysis}
To test the effectiveness of the spatiotemporal-discretized risk analysis algorithm, we design various hazardous scenarios, which are tested using both the risk field method and the traditional Time-To-Collision (TTC) algorithm. The accuracy of the risk field method is 95.83\%, whereas the traditional TTC achieves an accuracy of 62.50\%. The result of the risk field method is approximately 53.33\% better relative to the traditional TTC, with an average processing time of 26ms. Due to the superiority of the risk field model, the vehicle can perceive risks in advance and decelerate accordingly, avoiding the need for braking actions triggered by the TTC.

\subsubsection{Safe Trajectory Planning}
Numerous simulations are conducted to confirm that under normal traffic flow conditions, left turns, straight driving, right turns, and U-turns are effectively executed by our safe trajectory planning algorithm, ensuring safety and comfort with a traffic efficiency of no less than 90\% compared to human drivers.

\begin{figure}[htbp]
    \centering
    \subfigure[]{
        \includegraphics[width=0.4\textwidth]{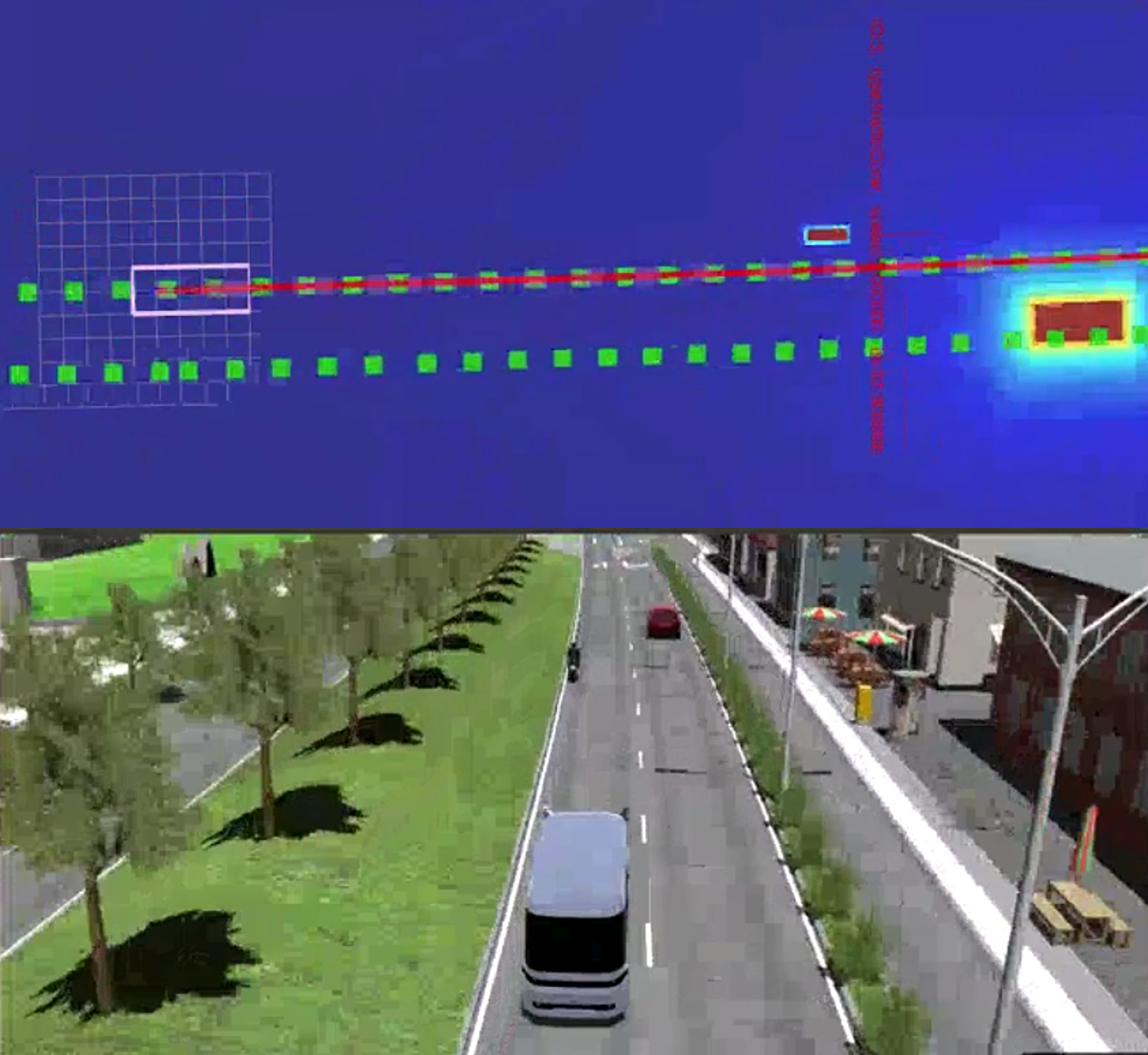}
    }
    \hfill
    \subfigure[]{
        \includegraphics[width=0.4\textwidth]{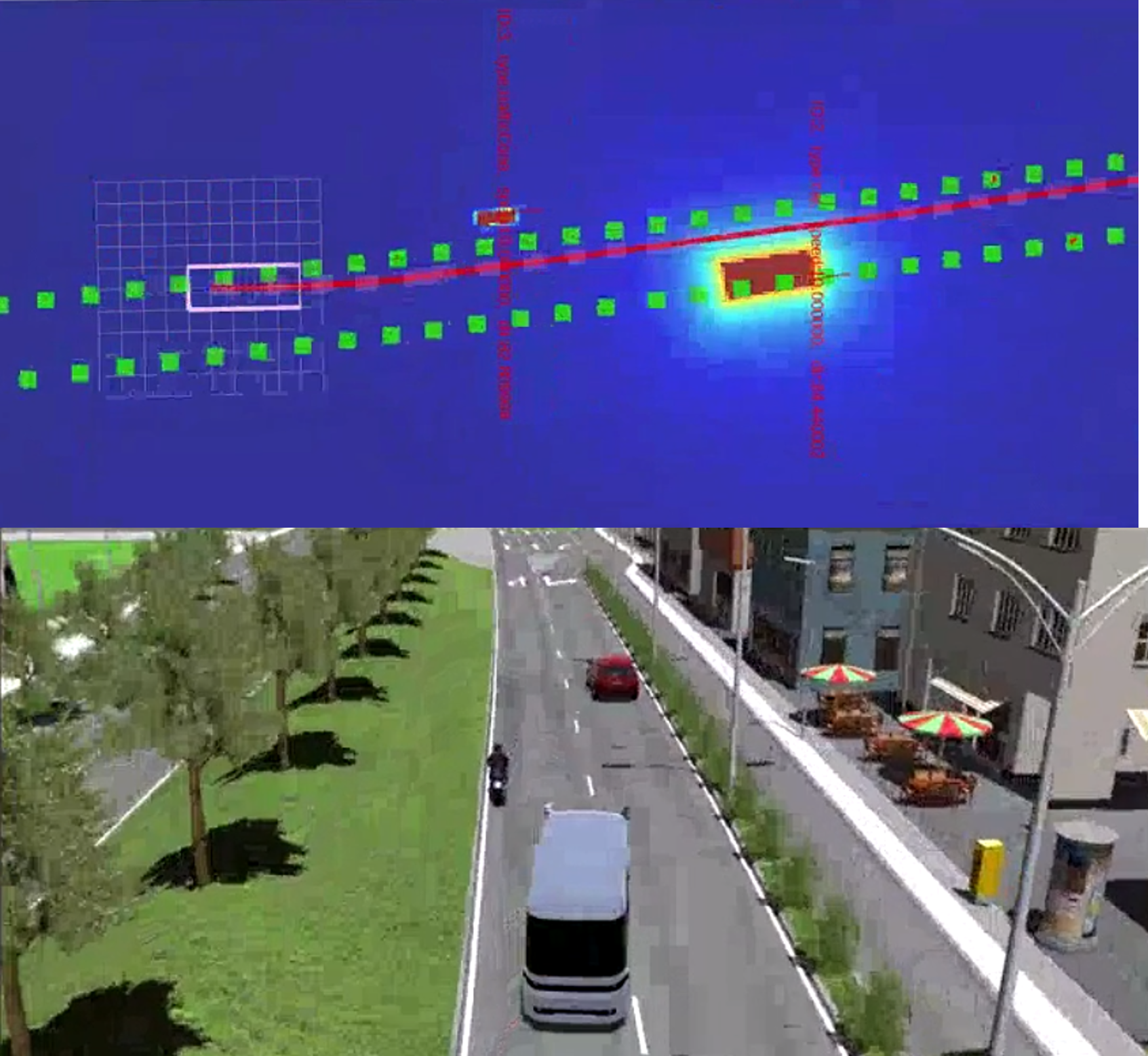}
    }
    \vfill
    \subfigure[]{
        \includegraphics[width=0.4\textwidth]{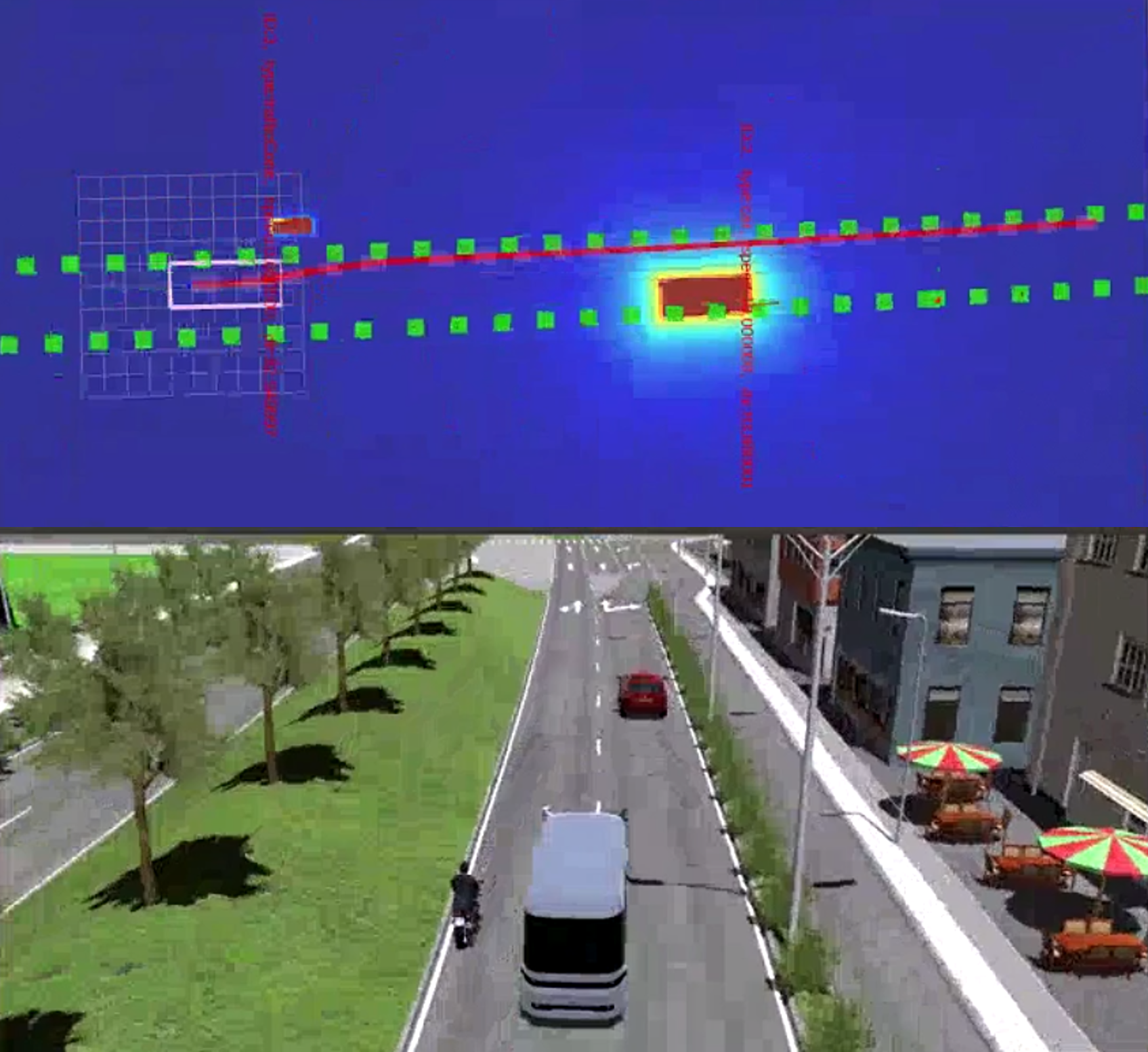}
    }
    \hfill
    \subfigure[]{
        \includegraphics[width=0.4\textwidth]{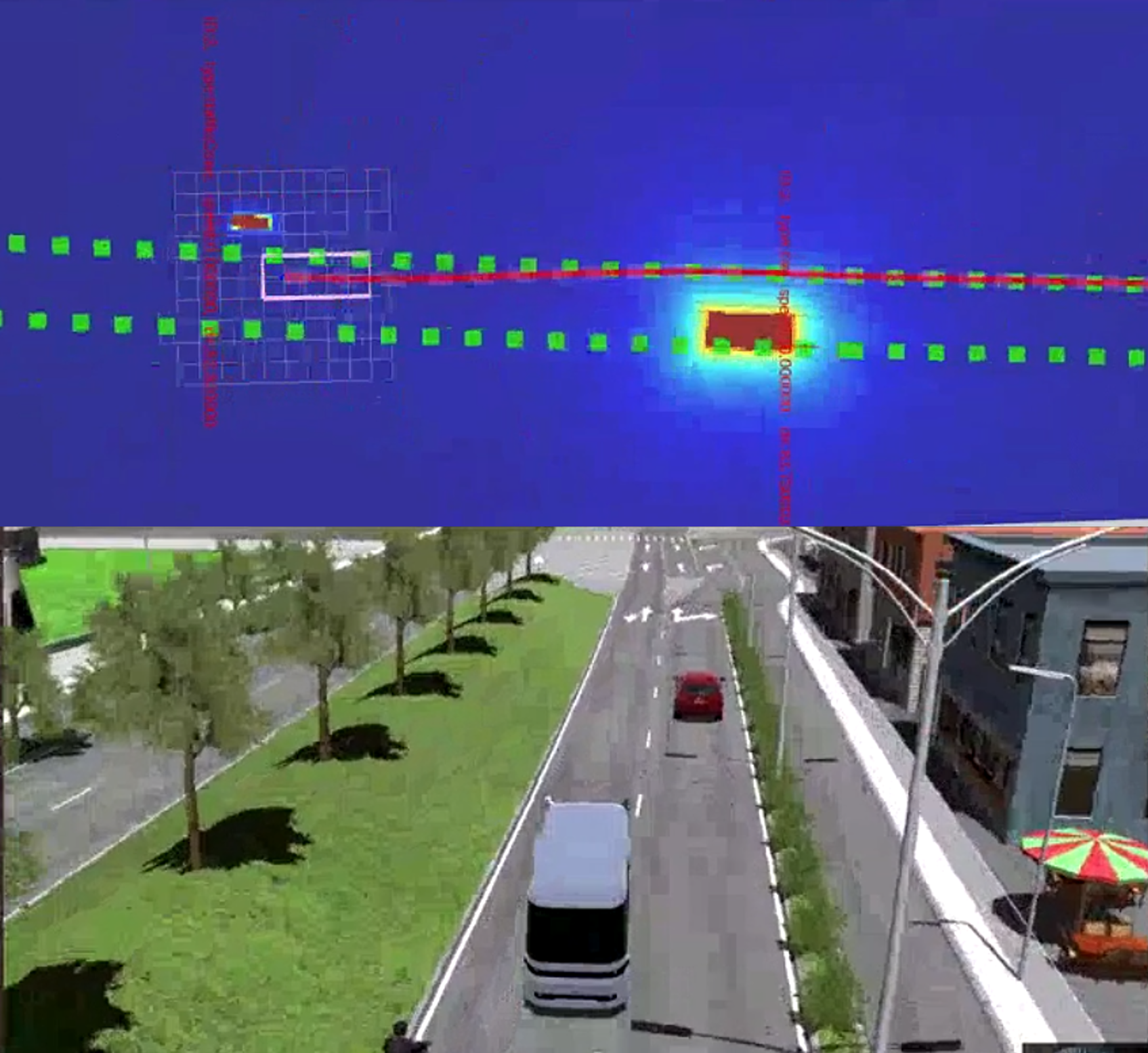}
    }
    \caption{An overtaking decision made based on risk field information when encountering slow-moving pedestrians by the roadside.}
    \label{Fig:Planning}
\end{figure}

Safety risks are accurately identified, and appropriate collision avoidance measures are taken; decisions are made based on environmental information, reflecting the vehicle's intelligence, avoiding unnecessary deceleration or stopping to ensure traffic efficiency. As shown in the simulation scenario in Fig. \ref{Fig:Planning}, an overtaking decision is made based on risk field information when encountering slow-moving pedestrians by the roadside.

Besides, our algorithm also achieve comfort planning with smooth linear acceleration and braking in non-emergency situations, with a maximum longitudinal acceleration of $1.07 m/s^2 (\leq 1.5 m/s^2)$ and a maximum lateral acceleration of $1.03 m/s^2 (\leq 1.2 m/s^2)$; safety is maintained by keeping a reasonable distance from obstacles.

\subsection{Experiment Results}

Before vehicle validation, considering the significant increase in computational load when calculating risk fields across multiple spacetimes, the computation functions of the spatiotemporal-discretized risk field are transferred to the GPU and utilized CUDA for multi-core parallel processing. The computed results are then transferred back to the CPU for further application of the risk field. Employing CUDA allows for real-time calculation of risk field information, ensuring the high real-time performance requirements needed for ICVs.

\begin{figure}[htbp]
    \centering
    \subfigure[The scene of a large truck in adjacent lane]{
        \includegraphics[width=0.58\textwidth]{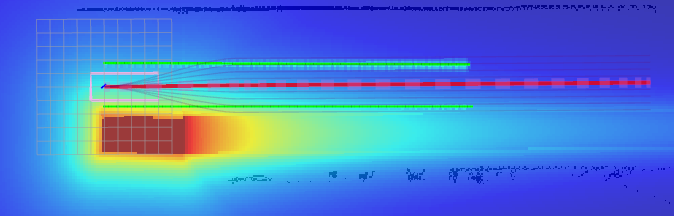}
    }
    \hfill
    \subfigure[Speed reduction in planned trajectory]{
        \includegraphics[width=0.36\textwidth]{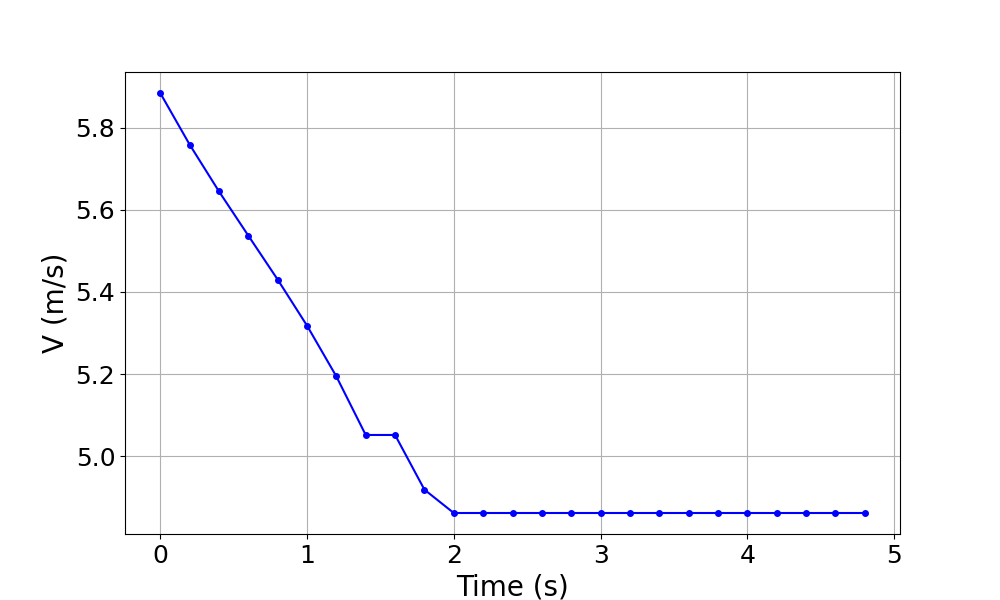}
    }
    \caption{The speed reduction when a large truck is in adjacent lane.}
    \label{Fig:validation}
\end{figure}

Vehicle experiment results illustrate that our whole algorithm achieves safe and comfortable trajectory planning for ICVs. Particularly, as Fig.\ref{Fig:validation} shows, when large trucks are in adjacent lanes, a slight reduction in speed is made to ensure safety. 

Furthermore, the time consumption of each module in the vehicle experiment is listed in Table. \ref{Tab:time}. As the three modules can operate in parallel, the real-time performance can be ensured for practical applications.

\begin{table*}[!htbp]
    \centering
    \caption{The time consumption of vehicle experiment.}
    \begin{tabular}{cc}
        \toprule
        Modules &Average Time Consumption(ms)\\
        \midrule
        Trajectory Prediction &65.96 \\
        Risk Analysis &26.00 \\
        Trajectory Planning &20.05 \\
        \bottomrule
    \end{tabular}
    \label{Tab:time}
\end{table*}

\section{Conclusion}\label{Sec:Conclusion}
The paper proposes a predictive risk analysis and safe trajectory planning framework for ICVs. A local risk-aware trajectory prediction algorithm is first employed to predict the future trajectories of surround objects, then spatiotemporal-discretised predictive driving risk is analyzed based on the prediction information. Finally, our approach generates safe trajectory of ego-vehicle by the predictive risk analysis. To validate the effectiveness and real-time performance of our approach, extensive simulation and vehicle experiments are conducted.

\backmatter

\bibliography{sn-bibliography}

\end{document}